\documentclass{vgtc}                          

\graphicspath{{figures/}{pictures/}{images/}{./}} 

\usepackage{times}                     

\usepackage{tabu}                      
\usepackage{booktabs}                  
\usepackage{lipsum}                    
\usepackage{mwe}                       

\usepackage{mathptmx}                  
\usepackage{abbrev}
\usepackage{amsmath}
\usepackage{amsfonts}

\onlineid{1241}

\vgtccategory{Technical papers}

\vgtcinsertpkg

\title{GO-NeRF: \underline{G}enerating \underline{O}bjects in \underline{Ne}ural \underline{R}adiance \underline{F}ields \\for Virtual Reality Content Creation}

\author{Peng Dai$^{1}$
\hspace{1cm} Feitong Tan$^{2}$ \hspace{1cm} Xin Yu$^{1}$ \hspace{1cm} Yifan Peng$^{1}$ \hspace{1cm} Yinda Zhang$^{2}$ \hspace{1cm} Xiaojuan Qi$^{1}$\\
 $^1$The University of Hong Kong \hspace{1cm} $^2$ Google\\
 \\
 Project page: \url{https://daipengwa.github.io/GO-NeRF/}
 }

\teaser{
  \centering
  \includegraphics[width=\linewidth]{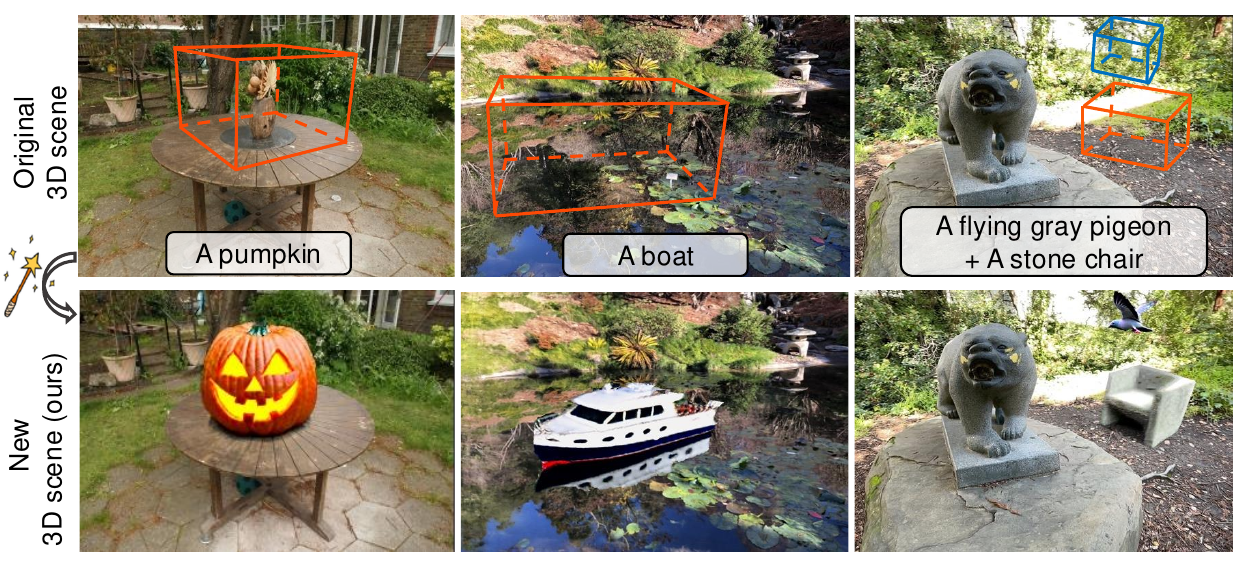}
  \vspace{-12pt}
  \caption{The capacity to generate new objects in an established 3D scene is fundamental for the creation and editing of virtual environments. Given 3D bounding boxes and textual prompts that describe virtual objects (top row), our approach focuses on generating virtual objects directly within pre-trained scene neural radiance fields, ensuring alignment with the 3D scene (bottom row). The last column reveals that our method also supports the generation of multiple objects.}
  \label{fig:teaser}
}

\abstract{
Virtual environments (VEs) are pivotal for virtual, augmented, and mixed reality systems. Despite advances in 3D generation and reconstruction, the direct creation of 3D objects within an established 3D scene (represented as NeRF) for novel VE creation remains a relatively unexplored domain. 
This process is complex, requiring not only the generation of high-quality 3D objects but also their seamless integration into the existing scene.
To this end, we propose a novel pipeline featuring an intuitive interface, dubbed \emph{GO-NeRF}. 
Our approach takes text prompts and user-specified regions as inputs and leverages the scene context to generate 3D objects within the scene. We employ a compositional rendering formulation that effectively integrates the generated 3D objects into the scene, utilizing optimized 3D-aware opacity maps to avoid unintended modifications to the original scene.
Furthermore, we develop tailored optimization objectives and training strategies to enhance the model’s ability to capture scene context and mitigate artifacts, such as floaters, that may occur while optimizing 3D objects within the scene. 
Extensive experiments conducted on both forward-facing and $360^o$ scenes demonstrate the superior performance of our proposed method in generating objects that harmonize with surrounding scenes and synthesizing high-quality novel view images. We are committed to making our code publicly available.}

\begin{document}


\firstsection{Introduction}

\maketitle


In recent years, significant progress has been made for re-renderable real-world environment reconstruction using neural radiance field (NeRF)~\cite{mildenhall2021nerf, barron2022mip, barron2023zip, barron2021mip, park2021nerfies, xu2022point, yu2021pixelnerf, fridovich2022plenoxels}. Concurrently, text-guided object generation~\cite{poole2022dreamfusion, wang2023prolificdreamer, yu2023text, liu2022iss, jain2022zero, liu2023zero1to3} has shown great promise in creating novel 3D contents. 
In this work, we explore a new problem: generating 3D objects directly within an established 3D scene to create a novel virtual environment. This is an important yet challenging problem as it demands the high-quality composition of generated content in the environment to ensure an immersive experience for downstream extended reality applications.

In practice, virtual environments are often created using computer graphics software like Blender, which requires skilled specialists to manually design scene layouts, geometries, materials, and rendering algorithms. While this approach can produce impressive results, the workflow and software operations are often tedious and complex. The advent of efficient reconstruction techniques has simplified some manual processes by allowing the use of reconstructed contents~\cite{zhang2021nerfactor, mildenhall2021nerf, wang2023neural}. However, real-world reconstruction is not always satisfactory. For example, the reconstructed 3D scenes (see~\cref{fig:teaser}, first row) may lack key objects (see~\cref{fig:teaser}, second row), emphasizing the need for re-creation capabilities. 

To address this, Gordon {\etal}~\cite{gordon2023blended} introduced a 3D blending pipeline for compositing independently synthesized 3D objects into established 3D scenes. However, this approach is limited by the model's generative capacity and its inability to leverage scene context, resulting in suboptimal, low-quality outcomes that fail to harmonize with the 3D scene (see \cref{fig:comparison}, second row: the fruits appear to be floating in the air). 
On the other hand, text-guided image inpainting models~\cite{Rombach_2022_CVPR, saharia2022photorealistic, stacchio2023stableinpainting} are trained to recreate masked regions with desired objects while utilizing the known scene context. Although these inpainted objects blend well with surrounding regions in 2D images, generating view-consistent images of the desired object for subsequent 2D-to-3D NeRF training~\cite{mirzaei2023spin, mirzaei2023reference} remains a challenge. As a result, these techniques are prone to large view inconsistencies and unintended scene modifications due to inaccurate inpainting masks (see \cref{fig:comparison}, bottom right: the provided mask does not align with the object's silhouette, leading to undesired alterations).

This work presents a novel pipeline featuring a user-friendly interface dubbed \emph{GO-NeRF}, which generates text-prompt-controlled 3D virtual objects at user-specified locations within an existing 3D environment, resulting in a harmonized new 3D scene (see \cref{fig:comparison} and \cref{fig:ab_all} (a) for examples of varying cat appearances with realistic poses and shapes across different scenes).
Our approach is underpinned by three key components: (1) an intuitive interface that allows users to specify areas within a 3D scene with just three clicks to generate virtual objects; (2) a compositional rendering formulation that seamlessly integrates the generated 3D objects into the scene while preventing unintended alterations; and (3) meticulously designed context-aware learning objectives to optimize the 3D objects, ensuring high quality and smooth fusion with the scene.

\begin{figure*}[h]
    \centering
    \includegraphics[width=1.0\linewidth]{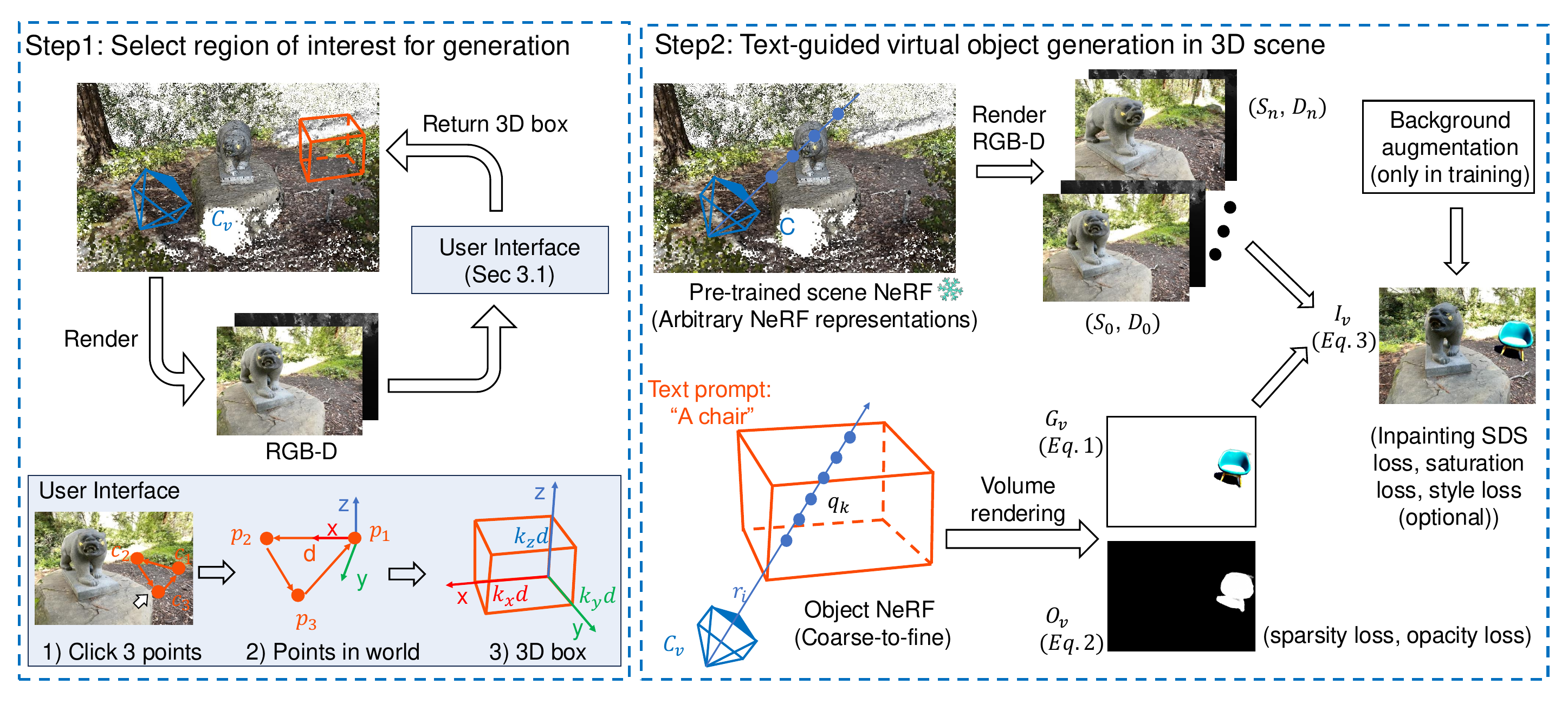}
    \vspace{-0.25in}
    \caption{\textbf{Virtual objects generation pipeline.} Left: we offer a user-friendly interface for specifying generation regions in the pre-trained 3D scene. Specifically, users can effortlessly define a 3D bounding box by selecting three points on the image. This is achieved by employing perspective projection and cross-product operations. Right: our approach separates scene rendering (up) and object generation (down) processes, which are subsequently combined in the rendered image space. The scene rendering phase generates RGB-D images $(S, D)$ of the 3D scene using pre-defined cameras $C$. The object generation step optimizes a neural radiance field within the 3D box to produce RGB images $G_v$ (Eq.~\eqref{equ:volume_rendering}) and opacity maps $O_v$ (Eq.~\eqref{equ:opacity}) through volume rendering techniques. Subsequently, the final output $I_v$ (Eq.~\eqref{equ:deferred_rendering}) is created by blending the scene and generated content using optimized opacity maps. Throughout the optimization, we meticulously design loss functions and training strategies to ensure the delivery of high-quality composited results.}
    \label{fig:pipeline}
\end{figure*}

Specifically, given a 3D scene, our interface allows users to select the 3D location for object generation by choosing three points from a rendered image. We then use depth information to convert these points into a 3D box within the scene (see \cref{fig:pipeline}, left). Within this specified 3D box, we create a new NeRF representation for the object, rendering it separately from the existing 3D scene.
Given a camera view, the rendered images of the scene and the object are composited using a 3D-aware opacity map to handle occlusions. This separation and 3D-aware composition preserve the original scene content outside the desired editing area, effectively manage occlusions, and ensure compatibility with established 3D scenes of various representations (e.g., InstantNGP~\cite{muller2022instant} and NeRF~\cite{mildenhall2021nerf}).

To optimize the object's NeRF based on a textual description and ensure its compatibility with the scene context, we distill 2D text-guided image inpainting priors from diffusion models~\cite{Rombach_2022_CVPR} using score distillation sampling (SDS)~\cite{poole2022dreamfusion}. This advanced inpainting prior allows us to effectively leverage scene context, facilitating the synthesis of scene-compatible objects. However, SDS-based results can suffer from oversaturation issues (see \cref{fig:ab_all} (b)). To address this, we introduce a regularizer that aligns the saturation of synthesized objects with the overall tone of the scene. Additionally, to eliminate artifacts that may arise during the object optimization process (see \cref{fig:ab_all} (a), where artifacts within the 3D box share a similar color with the scene background), we randomly replace the scene context with a random background at a certain ratio. Finally, we propose a reference-image-guided feature space loss, which uses feature space similarity to guide the style of the generated objects (see \cref{fig:more_application} (c)).
To demonstrate the effectiveness of our proposed method, we conduct extensive experiments on public datasets that include both forward-facing and $360^{\circ}$ scenes~\cite{barron2022mip, mildenhall2019llff, haque2023instruct}, showing superior performance in both quantitative and qualitative evaluations.  
In summary, our technical contributions are as follows:

\begin{itemize}
    \item We introduce GO-NeRF, a novel pipeline featuring a user-friendly interface that generates context-compatible 3D virtual objects from text prompts at user-specified locations within an established 3D scene, while preserving unchanged scene content and maintaining compatibility with various NeRF representations.
   
    \item We develop learning objectives and regularizers, enabling high-quality, floater-free 3D synthesis and composition to create new 3D virtual environments. 

    \item Experimental results showcase our approach outperforming previous methods on both forward-facing and $360^{\circ}$ datasets.
\end{itemize}

\section{Related Work}
\label{sec:related_work}
\noindent \textbf{Neural radiance field editing.} NeRF~\cite{mildenhall2021nerf} is primarily designed for novel view synthesis and has attracted significant attention due to its efficacy in reproducing our world~\cite{barron2021mip, xu2022point, dai2023hybrid, fridovich2022plenoxels, muller2022instant}. Considering the editing demands in NeRF, several works~\cite{wang2022clip, liu2021editing, haque2023instruct, zhuang2023dreameditor, chen2023gaussianeditor} attempted to modify the appearance and geometry of NeRF using diffusion priors or latent codes extracted from text prompts or RGB images. Huang {\etal}~\cite{huang2022stylizednerf} proposed to stylize NeRF using pre-stylized 2D images for NeRF fine-tuning; Kobayashi {\etal}~\cite{kobayashi2022decomposing} made the editing semantic-driven by distilling semantic features into NeRF; and Haque {\etal}~\cite{haque2023instruct} realized instruction-based NeRF editing by using a fine-tuned instruction-based image editing model~\cite{brooks2022instructpix2pix} as the supervisor. 
Moreover, NeRFs can be distilled onto the surface of explicit 3D mesh~\cite{yang2022neumesh, yuan2022nerf, bai2023learning}, enabling interactive editing with people. 
Instead of editing the appearance and geometry of content already in NeRF, Mirzaei {\etal}~\cite{mirzaei2023reference, mirzaei2023spin} proposed the inpainting NeRF task, which substituted regions of interest (ROI) with new content. Specifically, they adopted a 2D image inpainting model~\cite{suvorov2021resolution} to fill masked regions. Subsequently, the inpainted content was distilled into NeRF by using inpainted images for fine-tuning. Although impressive results have been delivered, their approach may struggle with scenes encountering drastic view changes. Unlike previous methods, we aim to generate virtual objects directly within the established scene's NeRF without being constrained by perspective changes.\\

\noindent \textbf{3D generation.} Accurately acquiring 3D content is valuable yet challenging for a diverse set of applications. Most ealier high-quality 3D models were crafted by experienced specialists; however, this creation process can be time-consuming. Recently, Dreamfuison~\cite{poole2022dreamfusion} proposed a text-guided 3D generation framework that optimized an object's NeRF using the novel Score Distillation Sampling (SDS) loss, which distills the generation capability of 2D diffusion model~\cite{Rombach_2022_CVPR} into 3D generation. This method greatly reduced consumption in high-quality 3D content creation and inspired numerous subsequent works~\cite{yu2023text, wang2023prolificdreamer, raj2023dreambooth3d, liu2023zero1to3, tang2023make, lin2023magic3d, kolotouros2023dreamhuman}. 
Unlike previous methods that focused on object-level generation, Giraffe~\cite{Niemeyer2020GIRAFFE} used compositional rendering to generate scenes containing multiple objects.
Similarly, Ryan {\etal}~\cite{po2023compositional} simultaneously optimized multiple pre-defined 3D bounding boxes with different text prompts and generated a scene. However, these methods are limited to single-category object generation or not conditioned on existing scenes. More recently, Gordon {\etal}~\cite{gordon2023blended} proposed to blend the generated objects into established 3D scenes using a distance-based blending scheme. Even though they successfully generated objects within the 3D scenes, the results were unrealistic and did not harmonize with the surrounding scene. In this paper, we focus on generating high-quality virtual objects that are coordinated with the established 3D scenes.

\section{Method}

\noindent \textbf{Overview.} We propose \emph{Go-NeRF}, a method designed to generate virtual objects within an established NeRF-based 3D scene based on a given text prompt. 
An overview of our pipeline is shown in \cref{fig:pipeline}. 
The initial step involves creating a 3D bounding box to identify the modification regions. This process is made easier by enabling users to select three points on the rendered images via our user-friendly interface (Sec.~\ref{sec: boudning_box}). Next, we introduce a compositional rendering pipeline (Sec.~\ref{sec: rendering_pipeline}) for the generation and integration of objects within the 3D scene. The core strategy involves decoupling object and scene rendering to enhance flexibility and merging them using a 3D-aware opacity map to handle occlusions effectively. 
we devise effective loss functions and training strategies (Sec.~\ref{sec: loss}) to direct the optimization process and enable the generation of objects that seamlessly integrate with the scene. 

\subsection{Interface}
\label{sec: boudning_box}




We develop a user-friendly interface that simplifies the process of positioning objects within a 3D scene for generation, catering to casual users. This tool eliminates the need for complex 3D user interfaces (e.g., Blender), allowing users to effortlessly define object positions by clicking on 2D images to automatically create corresponding 3D bounding boxes, as illustrated in \cref{fig:pipeline} left. 
The process begins with the selection of three points, denoted as $\{c_1, c_2, c_3\}$, which are then back-projected onto the 3D scene using their depth values, yielding points $\{p_1, p_2, p_3\}$ and enabling the construction of a plane $P$. Subsequently, the coordinate system for the 3D bounding box is established by defining the $x$ axis as the vector from $p_1$ to $p_2$, setting the $z$ axis perpendicular to plane $P$, and computing the $y$ axis through cross-product operations. The size of each 3D bounding box (i.e., the length of the box along each axis) is then manually determined as a ratio $\{k_x, k_y, k_z\}$ of the distance $d$ between $p_1$ and $p_2$. To position objects suspended in free space, an upward movement along the $z$ axis is applied to elevate the object to the desired free space location.


It is worth noting that the 3D bounding box does not necessarily have to align with the object's shape, contrary to the requirements of 2D inpainting-based methods~\cite{mirzaei2023reference}, where an accurate object mask is essential to avoid unintended modifications. Instead, we view the 3D box as a rough constraint on the object's size and dynamically optimize occupancy values within the box to generate the object.


\subsection{Compositional Rendering}
\label{sec: rendering_pipeline}

As illustrated in the right panel of \cref{fig:pipeline}, in addition to the initial scene representation $F(\theta_{s})$, where $\theta_{s}$ is optimized based on the original scene images, a distinct NeRF $F(\theta_o)$ is introduced to model the object, parameterized by $\theta_o$, which will be optimized during training. Then, given a camera viewpoint $v$, the scene image $S_v$ and object image $O_v$  are rendered separately. These rendered images are then combined together to produce the final rendered output $I_v$. We elaborate on the image rendering below.

The rendering of the scene image $S_v$ is a straightforward process, where we generate RGB-D images from the established scene's NeRF $F(\theta_{s})$ using its rendering formula. Given that the scene content remains constant while optimizing $F(\theta_o)$, we seek to reduce the computational load of repeated volume rendering by pre-rendering the established 3D scene into a series of RGB-D image sequences $\{(S_0, D_0), ..., (S_{n}, D_{n})\}$ from a predefined set of camera viewpoints $\{C_0, ..., C_{n}\}$. In practice, we employ the same camera viewpoints used for training the scene's NeRF $F(\theta_{s})$.

Subsequently, for object rendering, RGB images $\{G_0, ..., G_{n}\}$ and opacity maps $\{O_0, ..., O_{n}\}$ are rendered from the object's NeRF $F(\theta_o)$ within the 3D bounding box. This involves casting rays from a specific camera viewpoint $v$ ($v \in \{C_0, C_1,..., C_n\}$ during optimization) and sampling query points inside the 3D box. The RGB value $G_v(i)$ and opacity value $O_v(i)$ of a ray $r_i$ are then rendered based on following equations: 
\vspace{-6pt}
\begin{equation}
\small
\begin{aligned}
& \tau_k = \text{exp}(-\sum_{t=1}^{k-1}\delta_t\Delta_t),\\
    G_v(i) &= \sum_{k=1}^{K}\tau_k(1-\text{exp}(-\delta_k\Delta_k))c_k;\\
\end{aligned}
    \label{equ:volume_rendering}
\end{equation}
\begin{equation}
    \small
    O_v (i) = \left \{
    \begin{aligned}
        &\sum_{k=1}^{K}\tau_k(1-\text{exp}(-\delta_k\Delta_k)), \\
        &0, \quad if \quad r_i \notin \text{box} | D_\text{box}(i) > D_v(i).
    \end{aligned}
    \right.
    \label{equ:opacity}
\end{equation}
$\delta_{k}$ and $c_{k}$ are volume density and RGB values at query point $q_{k}$, and $\Delta_{k}$ represents the distance between two adjacent query points
along the ray. Additionally, we use the original scene content ($S_v$ in Eq.~\eqref{equ:deferred_rendering}) by setting the opacity value as 0, when the ray has no intersection with the 3D box ($ r_i \notin \text{box}$) or the generated content is occluded by the scene foreground ($D_\text{box}(i) > D_v(i)$).     

To obtain the final output $I_v$ for viewpoint $v$, we leverage the opacity map $O_v$ to composite the rendered object $O_v$ and scene $S_v$ in image space following Eq.~\eqref{equ:deferred_rendering}: 
\begin{equation}
    \small
    I_v = G_v\cdot O_v + S_v\cdot (1 - O_v).
    \label{equ:deferred_rendering}
\end{equation}

The opacity map, obtained after optimizing $F(\theta_o)$, precisely delineates the regions corresponding to the generated objects (shown as white regions in \cref{fig:pipeline} bottom right). Utilizing this map to guide the composition process is beneficial for preserving the unchanged content of the scene. Additionally, although the composition occurs in the image space, occlusions are effectively handled by comparing the depth values of the scene and the object in 3D space as shown in Eq.~\eqref{equ:opacity}. Finally, the compositional rendering also ensures that our generation pipeline is unaffected by the methods used for scene pre-training, making it a plug-and-play solution compatible with various 3D scene representations.

\subsection{Optimization}
\label{sec: loss}
In the following, we first describe the losses used for object generation and then introduce optimization strategies to enhance the quality of the generated objects. For simplicity, we will omit the subscript $v$ from $\{I_v, G_v, O_v, S_v\}$ in the discussion of the loss functions, unless otherwise specified.\\

\noindent \textbf{Inpainting SDS loss.} 
We employ a pre-trained 2D diffusion model, denoted as $\epsilon_{\phi}$, to provide generative priors that guide the optimization of the 3D object $F(\theta_o)$. Our approach involves optimizing $F(\theta_o)$ by supervising rendered 2D images $I$ through score distillation sampling (SDS). However, directly using a text-to-image diffusion model, similar to DreamFusion \cite{poole2022dreamfusion}, does not guarantee alignment between the generated object and the existing 3D scene. Drawing inspiration from 2D image inpainting, which generates content conditioned on known regions of images, we propose employing a diffusion-based inpainting model~\cite{Rombach_2022_CVPR} for score distillation.
Specifically, given a mask $M$ and a masked image $I_M$, our SDS loss, derived from the inpainting-based diffusion model, is defined as follows:
\begin{equation}
    \begin{aligned}
        \nabla_{\Theta} \mathcal{L}_{\text{SDS}} & = \mathbb{E}_{t,\epsilon,C} \left[ \omega(t) \cdot \left[ \epsilon_{\phi}(I_t; y, I_M, M, t) - \epsilon \right] \cdot \frac{\partial I}{\partial \Theta} \right], 
    \end{aligned}
\end{equation}
where $t$ represents a time step randomly sampled within the diffusion process,  $ \epsilon $ is a randomly sampled Gaussian noise, $I_t$ is the noise-perturbed image, and the mask $M$ is derived by projecting the 3D box into a camera viewpoint. 
This objective function encourages the rendered images to reside in high-density areas \cite{poole2022dreamfusion}, conditioned on both the text prompt $y$ and the scene information provided by $M$ and $I_M$. As a result, it ensures a high-quality, harmonized composition of the optimized object within the established scene. \\

\noindent \textbf{Geometry loss.}
Following~\cite{jain2022zero}, we employ sparsity loss and opacity loss to facilitate the optimization of the object's geometry. 1) The sparsity loss encourages the rendered opacity map to be sparse, as defined by Eq.~\eqref{equ: loss_sparsity}: 
\begin{equation}
    \begin{aligned}
        \mathcal{L}_{\text{S}} &= \frac{1}{N} \sum_{i=1}^{N} O(i),
    \end{aligned}
    \label{equ: loss_sparsity}
\end{equation}
where $N$ is the number of rays intersecting with the 3D box. This loss benefits compact object generation, effectively suppressing floaters. 2) The opacity loss aims to avoid translucent effects by encouraging opacity values to be 0 or 1 using Eq.~\eqref{equ: loss_opacity}:
\begin{equation}
        \mathcal{L}_{\text{O}} = -\frac{1}{N}\sum_{i=1}^{N} O(i) \cdot \log(O(i)) + (1 - O(i)) \cdot \log(1 - O(i)).\\
    \label{equ: loss_opacity}
\end{equation}

\noindent \textbf{Saturation loss.} 
While the SDS loss boosts the generation of the 3D object, it suffers from color over-saturation issues~\cite{wang2023prolificdreamer}, hindering the composition of the generated object into the established scene to produce a coherent scene. To mitigate this issue, we utilize the saturation values from the reference image to constrain the generated object, as defined by Eq.~\eqref{equ: loss_saturation}:
\begin{equation}
    \begin{aligned}
        \mathcal{L}_{\text{SAT}} &= \left( \overline{G}_{s} - \overline{R}_{s} \right)^{2} + \left( \hat{G}_{s} - \hat{R}_{s} \right)^{2},
    \end{aligned}
    \label{equ: loss_saturation}
\end{equation}
where $\overline{G}_{s}$ and $\hat{G}_{s}$ denote the mean and variance of the saturation values for the generated content, masked using the opacity map $O$. Similarly, $\overline{R}_{s}$ and $\hat{R}_{s}$ represent the mean and variance of saturation values for the reference image $R$. Unless explicitly specified, we employ the rendered scene image as the reference ($R := S$) for computing the saturation loss.\\

\noindent \textbf{Style loss.} 
 We further incorporate a style loss to improve the color and style coherence between generated objects and a given reference. Unlike the saturation loss, which solely focuses on addressing the over-saturation issues, this loss captures the feature-level information from the reference to constrain the generated object:
\begin{equation}
    \begin{aligned}
        \mathcal{L}_{\text{STY}}^{\text{global}} &= \left( \overline{G}_{\text{vgg}} - \overline{R}_{\text{vgg}} \right)^{2} + \left( \hat{G}_{\text{vgg}} - \hat{R}_{\text{vgg}} \right)^{2}, \\
        \mathcal{L}_{\text{STY}}^{\text{local}} &= \frac{1}{N} \sum_{i=1}^{N} \min_{i'} \left( G_{\text{vgg}}(i) - R_{\text{vgg}}(i') \right)^{2}, \\
        \mathcal{L}_{\text{STY}} &= \mathcal{L}_{\text{STY}}^{\text{global}} + \mathcal{L}_{\text{STY}}^{\text{local}}.
    \end{aligned}
    \label{equ: loss_style}
\end{equation}
This style loss $\mathcal{L}_{\text{STY}}$ in Eq.~\eqref{equ: loss_style} consists of $\mathcal{L}_{\text{STY}}^{\text{{global}}}$ and $\mathcal{L}_{\text{STY}}^{\text{local}}$. Similar to saturation loss, the $\mathcal{L}_{\text{STY}}^{\text{global}}$ calculates the statistical loss in VGG feature space~\cite{johnson2016perceptual}, and the $\mathcal{L}_{\text{STY}}^{\text{local}}$ is a contextual loss~\cite{mechrez2018contextual} that searches the closest feature for measuring the difference.\\

\noindent \textbf{Overall loss.} Finally, the overall loss Eq.~\eqref{equ: loss_func} is formulated as a weighted combination of all loss terms:
\begin{equation}
    \begin{aligned}
        \mathcal{L} &= \mathcal{L}_{\text{SDS}} + \lambda_{\text{S}} \cdot \mathcal{L}_{\text{S}} + \lambda_{\text{O}} \cdot \mathcal{L}_{\text{O}} + \lambda_{\text{R}} \cdot \mathcal{L}_{\text{SAT or STY}}.
    \end{aligned}
    \label{equ: loss_func}
\end{equation}
Note that the generated low-light regions are excluded using an empirically defined intensity threshold ($<0.2$) and are not used for calculating saturation or style loss.\\

\noindent \textbf{Coarse-to-fine optimization.}
Our object's NeRF adopts the hash grid representation for efficient 3D content generation~\cite{muller2022instant}. Instead of optimizing a high-resolution hash grid at the beginning, which tends to overfit training views, we start with a low-resolution hash grid and gradually increase its resolution to facilitate generating compact objects.\\

\begin{figure*}[h]
    \centering
    \includegraphics[width=0.97\linewidth]{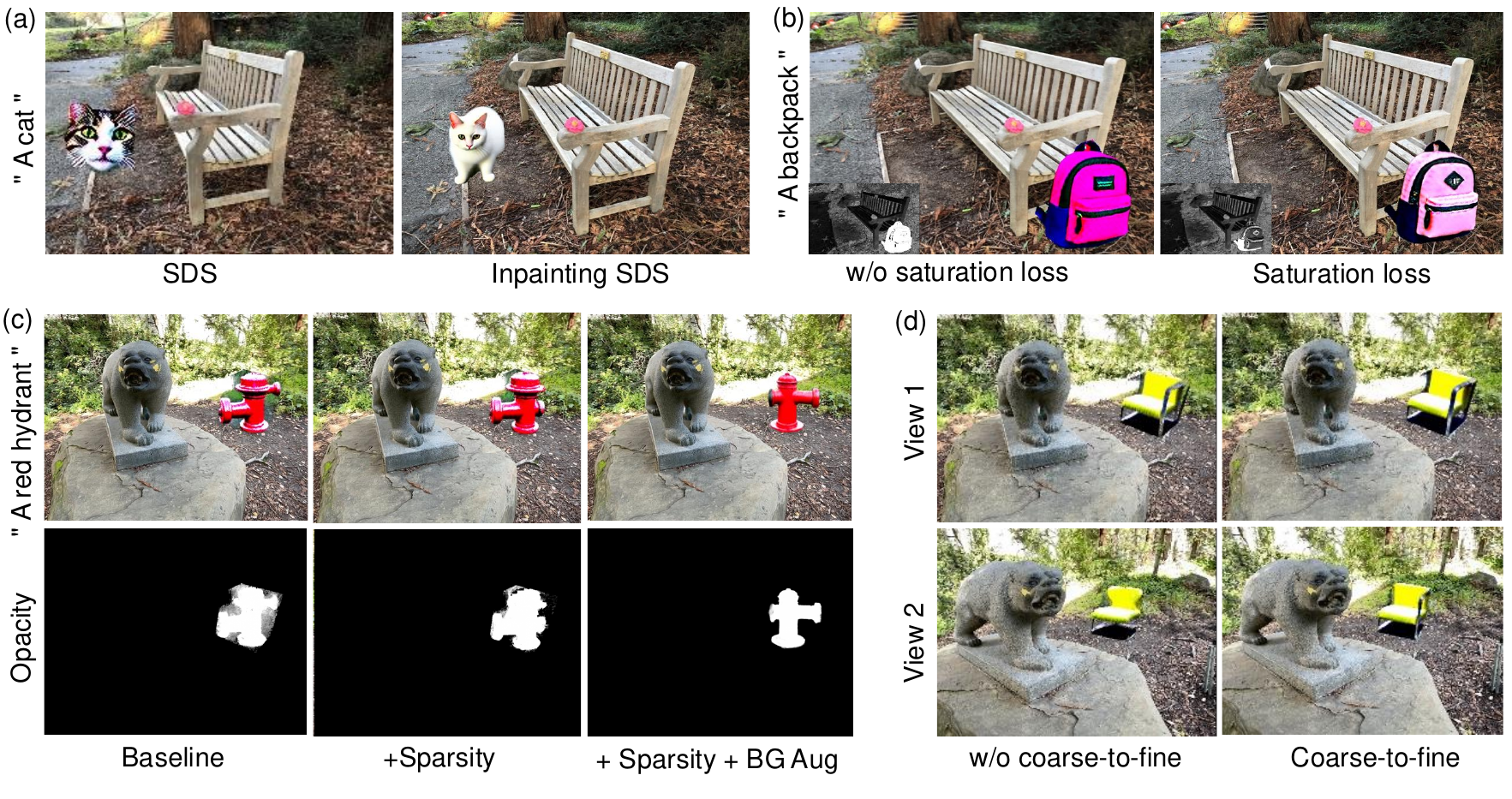}
    \vspace{-0.22in}
    \caption{\textbf{Ablation studies.} (a) Our proposed inpainting SDS loss effectively utilizes the scene context to generate a cat with an accurate shape and pose, whereas the standard SDS loss only produces a cat's head. (b) We present rendered RGB images alongside their corresponding saturation maps at the bottom left, where bright regions indicate high saturation values. Without constraining the saturation values, the generated backpack appears over-saturated. (c) The generated content is marred by artifacts that closely resemble the scene background, making their removal challenging. The opacity maps in the second row provide a clear visualization of this issue. The best results are achieved when employing both sparsity loss and background augmentation. (d) The coarse-to-fine optimization strategy improves the generation of compact and view-consistent objects, as exemplified by the chair's shape when viewed from different camera perspectives.}
    \label{fig:ab_all}
\end{figure*}

\noindent \textbf{Background augmentation.}
Since our method composites the generated content $G_v$ and established scene $S_v$ for optimization, the generated content often includes artifacts that closely resemble the scene background (see \cref{fig:ab_all} (c) first column). This similarity makes them difficult to observe and remove. To address this challenge, we augment the scene context with pure white or black during optimization, which makes the floaters more pronounced and easier to eliminate. It works in conjunction with the sparsity loss (Eq.~\eqref{equ: loss_sparsity}) to generate objects with a clean background (see \cref{fig:ab_all} (c) third column).

\section{Experiments}
\begin{figure*}
    \centering
    \includegraphics[width=1.0\linewidth]{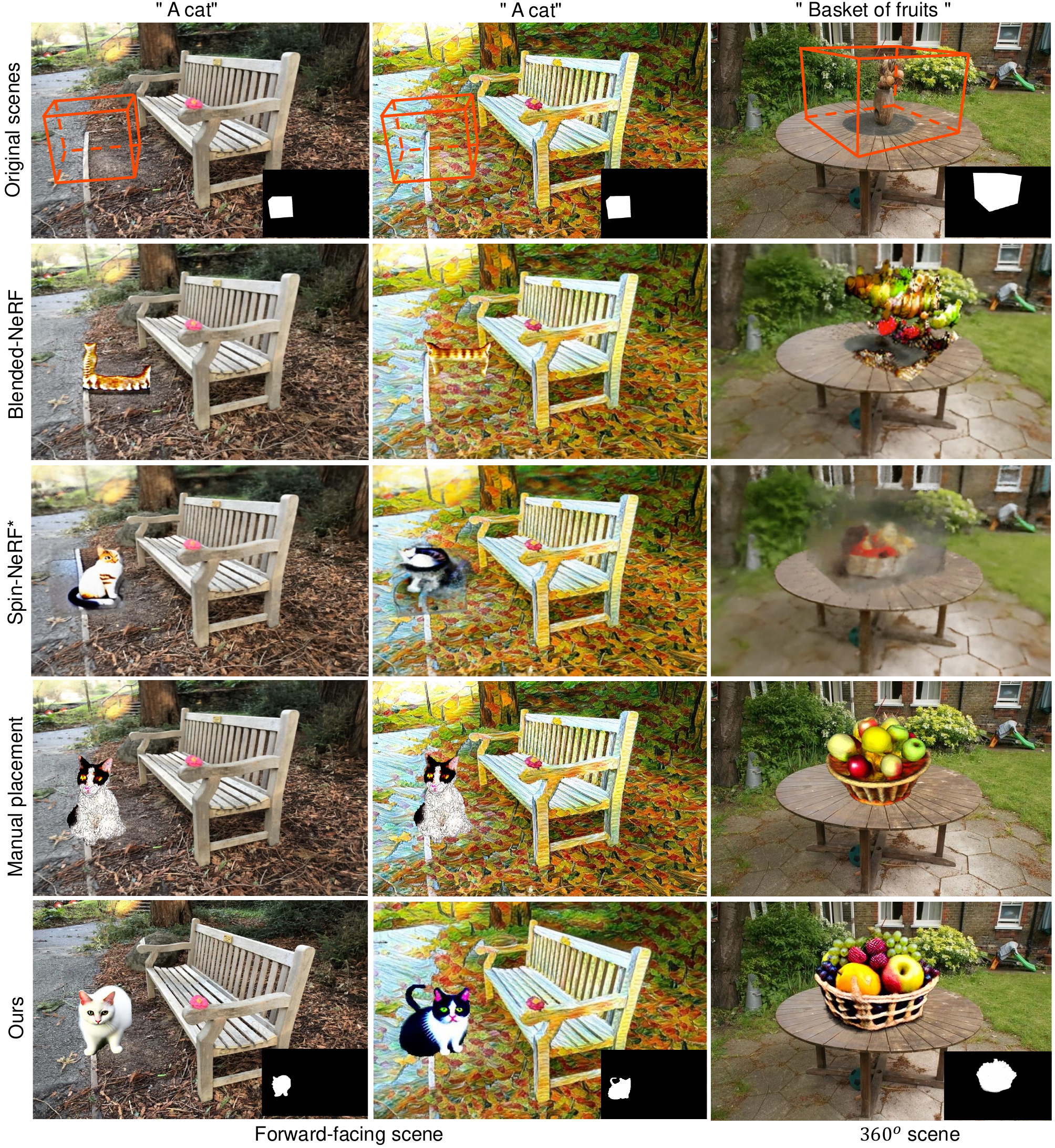}
    \vspace{-6pt}
    \caption{\textbf{Qualitative comparison}. We compare our method with other baselines on forward-facing and $360^o$ scenes. The first row displays the 3D box alongside its corresponding 2D mask in image space, while the subsequent rows present the results of various methods. Blended-NeRF tends to produce unrealistic and disharmonious results, such as fruits floating in the air. Spin-NeRF$^*$ failed in stylized scenes and $360^o$ scenes with large view changes. Moreover, manual placement is tedious and ignores the influence of scene context. In contrast, our method excels across all scenes, producing cats with different appearances and fruits on the table, accompanied by plausible shadows that enhance overall composition quality. At the bottom right of the last row, we visualize the optimized opacity maps that precisely describe the silhouette of generated content.}  
    \label{fig:comparison}
\end{figure*}



\begin{figure*}[htb]
    \centering
    \includegraphics[width=0.97\linewidth]{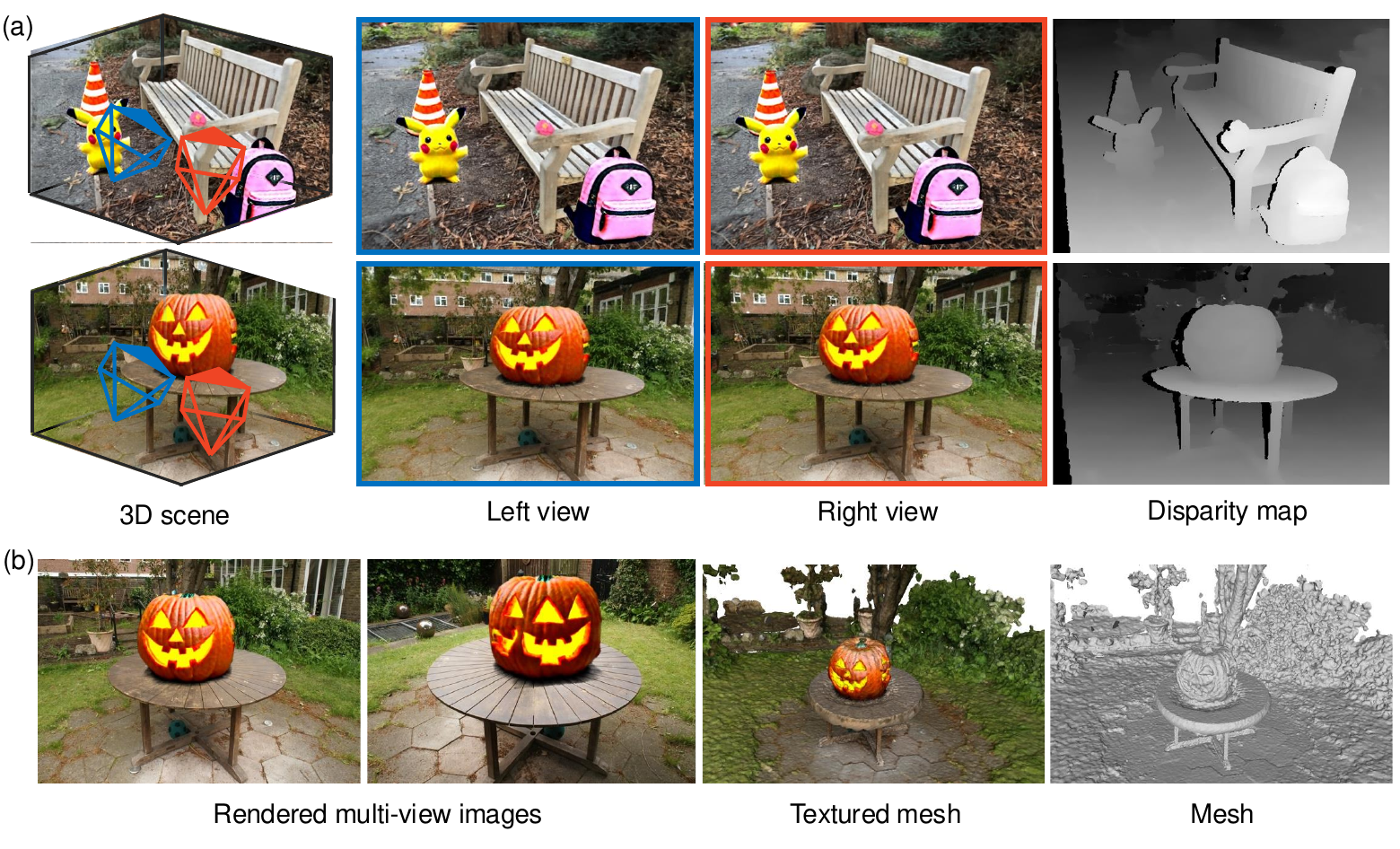}
    \vspace{-0.21in}
    \caption{\textbf{Stereoscopic results for VR \& 3D reconstruction.} (a) We render the new scene with generated 3D objects into stereoscopic results and visualize their stereo effects by predicting the disparity map from rendered left- and right-view images using stereo transformer~\cite{li2021revisiting}. The resulting disparity maps exhibit sharp details with clear foreground and background distinctions, indicating high-quality stereo effects. (b) We render the scene with generated virtual objects from different camera perspectives. Following the recent work~\cite{tancik2023nerfstudio}, we use those rendered multi-view images to refine NeRF optimization and extract the underlying 3D mesh. The successful 3D mesh reconstruction showcases consistency across diverse camera views.}
    \label{fig:more_stereo}
\end{figure*}


\begin{figure*}[htb]
    \centering
    \includegraphics[width=0.97\linewidth]{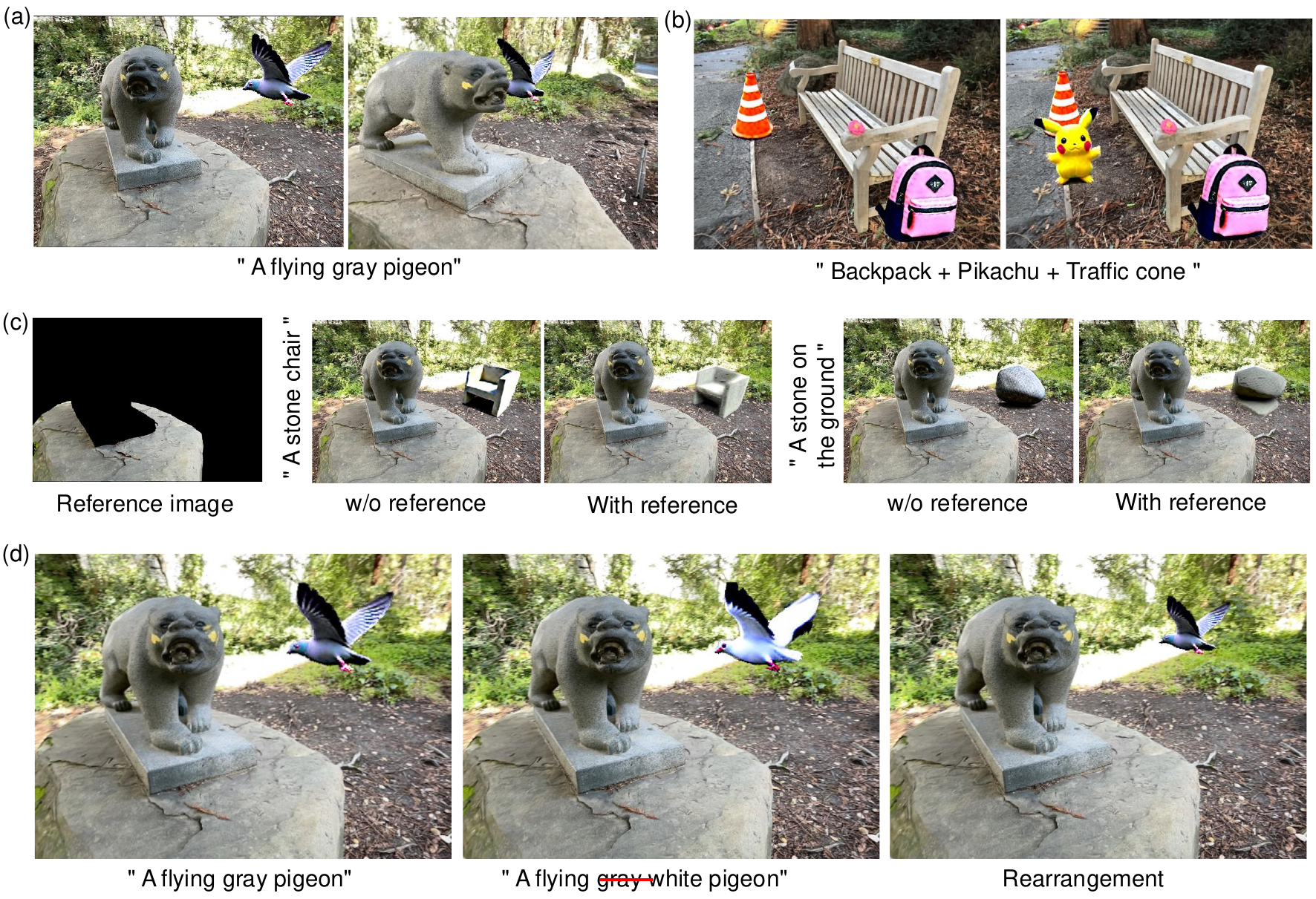}
    \vspace{-9pt}
    \caption{\textbf{Results of various application scenarios where our Go-NeRF framework aids.} (a) Our method successfully generates suspended objects, exemplified by a bird flying in the air, and captured by various  camera perspectives.
    (b) Our method facilitates generating multiple objects within an established 3D scene.
    (c) Style adaptation is seamlessly integrated. By utilizing a reference image as a guide, the generated object mirrors the visual characteristics of the reference image. 
    (d) Editing capabilities are robust. By adjusting the input text prompt, we can easily customize the appearance of generated objects, such as altering the bird's color. Furthermore, the decomposed representation of the scene and objects allows for effortless rearrangement of generated elements.}
    \label{fig:more_application}
\end{figure*}


\begin{figure*}
    \centering
    \includegraphics[width=0.97\linewidth]{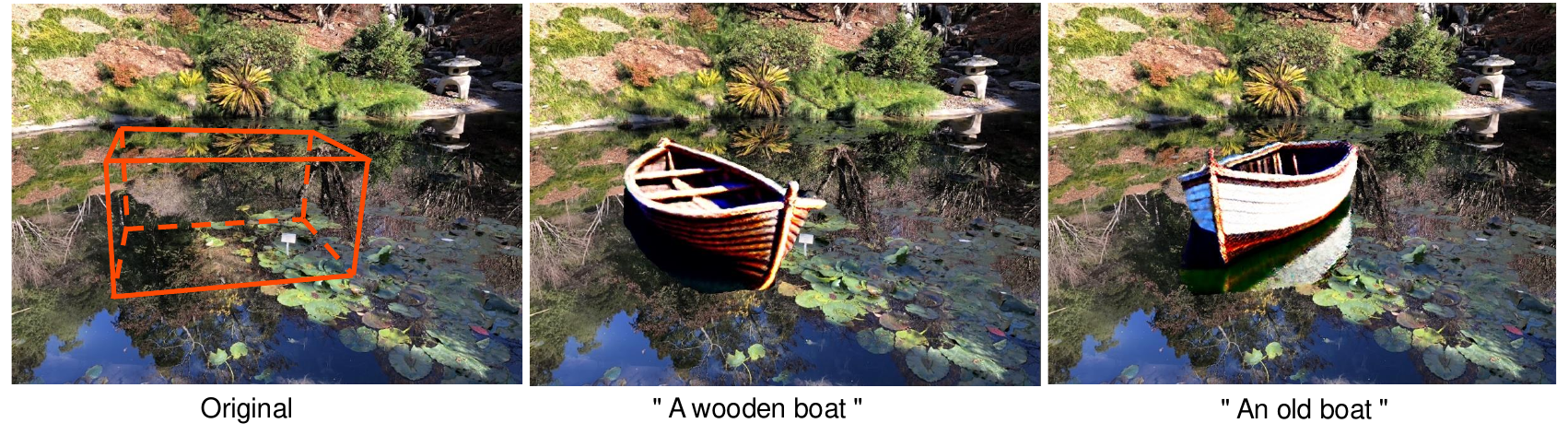}
    \vspace{-0.15in}
    \caption{\textbf{Results tested on a scene incorporating the reflective surface.} In our endeavor to create virtual boats navigating a complex water surface, we made an intriguing discovery: the generated boat exhibits reflections. Although these reflections are presently imperfect and may not align with physical correctness, this observation highlights the possibility of leveraging diffusion priors learned from extensive data to incorporate rendering-relevant effects in upcoming endeavors.}
    \label{fig:more_boat}
\end{figure*}

\begin{figure*}
    \centering
    \includegraphics[width=0.96\linewidth]{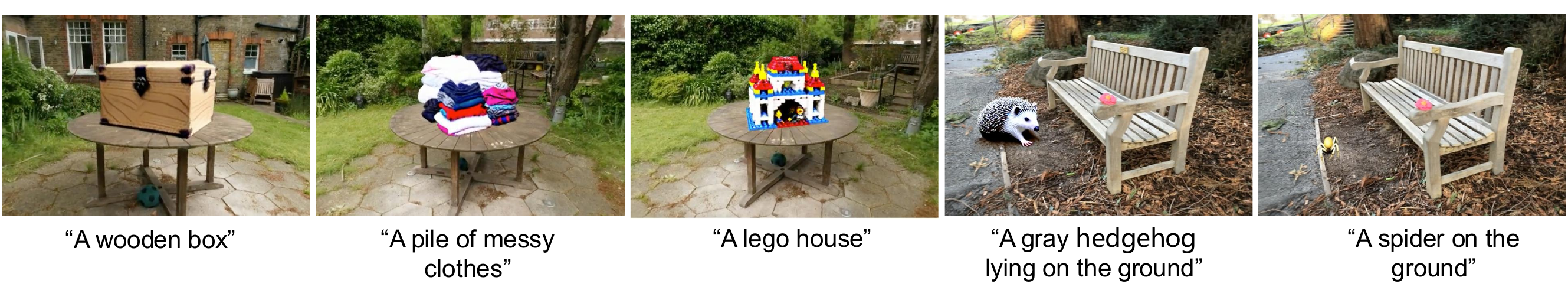}
    \vspace{-0.15in}
    \caption{\textbf{Additional results of scene editing with various VR content generated by the proposed Go-NeRF}.}
    \label{fig:more_cases}
    \vspace{-9pt}
\end{figure*}

\subsection{Implementation Details}
\noindent \textbf{Network and training.} To obtain established scene neural radiance fields and valid the compatibility of different NeRF representations, we use the PyTorch implementation of NeRF~\cite{lin2020nerfpytorch} for forward-facing scenes, and nerfstudio~\cite{tancik2023nerfstudio} for $360^o$ scenes. Following the 3D object generation pipeline~\cite{threestudio2023}, we optimize a hash grid representation~\cite{muller2022instant} within the 3D bounding box. During training,  $\lambda_S$ and $\lambda_O$ are determined using a cosine scheduler with values ranging from 30 to 300, and $\lambda_R$ is set as 500. We optimize our generation model on a single Nvidia 3090 GPU for a total of 20,000 iterations and $30\%$ iterations are trained with background augmentation.\\

\noindent \textbf{Datasets.} We conduct experiments on the publicly available datasets, including both forward-facing scenes and $360^o$ scenes from Instruct-NeRF2NeRF~\cite{haque2023instruct}, LLFF~\cite{mildenhall2019llff}, and Mip-NeRF $360^o$~\cite{barron2022mip}. The specific scenes we use are ``Bear", ``Garden", ``Benchflower", and ``Pond".\\

\noindent \textbf{Baselines.}
\emph{1) Manual placement.} We manually place the virtual object, generated using SDS loss~\cite{poole2022dreamfusion}, into the scene. \emph{2) Blended-NeRF}~\cite{gordon2023blended}. It optimizes an object's NeRF within a 3D bounding box using CLIP-based loss~\cite{radford2021learning}. Subsequently, the generated object is blended into the scene's NeRF based on its distance to the object center. Note that no scene context is used to guide optimization and blending. \emph{3) Spin-NeRF$^*$}~\cite{mirzaei2023spin}. To modify regions within the established scene, it employs a 2D image inpainting model, i.e., LaMa~\cite{suvorov2021resolution}, to fill masked regions in image space. Subsequently, the scene's NeRF is fine-tuned using these inpainted images. Here, we replace the LaMa~\cite{suvorov2021resolution} with the stable diffusion inpainting model~\cite{Rombach_2022_CVPR} to make this process text-guided and obtain multi-view inpainted images by gradually warping and filling the dis-occluded regions~\cite{mirzaei2023reference}.

\subsection{Qualitative Comparison}
We compare our method with baselines on both forward-facing and $360^o$ scenes. To validate that the scene context has an influence on the generation process, we additionally stylized the forward-facing scene into an oil painting style using ARF~\cite{zhang2022arf} but experimented with the same text prompt. The comparison results are displayed in \cref{fig:comparison}, where the first row visualizes the 3D bounding box and its corresponding mask region in the 2D image plane.

Without using the scene context for guidance, the generated objects in Blened-NeRF do not composite well with the scene (\cref{fig:comparison} second row). For instance, the fruits float in the air instead of resting on the table. Moreover, its overall quality remains low due to the limitations of using clip-based loss. In contrast, Spin-NeRF$^*$ utilizes scene context by employing a 2D inpainting model~\cite{Rombach_2022_CVPR} to fill the masked regions (first-row bottom right), leading to more reasonable results (\cref{fig:comparison} third row). However, this approach still exhibits several weaknesses: 1) the mask-based inpainting scheme inevitably alters scene content (e.g., \cref{fig:comparison} first column, the white line is extended.) due to misalignment between mask and object silhouette; 2) The rendering fidelity heavily relies on the quality of inpainted images which are affected by the accuracy of estimated depth~\cite{ranftl2021vision}. Thus, this method is unsuitable for scenes with significant viewpoint changes or those present challenges in depth estimation. This is evident in stylized and $360^o$ scenes, where the performance of Spin-NeRF$^*$ declines. Instead, generating virtual objects and then manually placing them into the scene is a tedious process requiring users to be proficient in 3D software operation. Moreover, this approach overlooks the impact of scene context, resulting in the same cat across different scenes (\cref{fig:comparison} fourth row). Additionally, extracting maneuverable mesh from the NeRF often degrades the fidelity of generated content.

Our method achieves good performance on both forward-facing and $360^o$ scenes. In \cref{fig:comparison} last row, the cat exhibits varying appearances in different scenes, where the cat resembles a painting in the stylized scene. The generated fruits are realistically placed on the table, and the plausible shadow around the base of the basket enhances the composition quality. When compared to manual placement, our results are more realistic. We suspect that the photo-realistic environment positively contributes to the realism of generated objects. Additionally, our approach composites generated objects into the scene employing optimized opacity maps (last-row bottom right), which accurately describe the silhouette of generated objects, thus preserving unchanged scene content. Please refer to the supplementary material for video results.

\subsection{Quantitative Comparison}
We use CLIP score~\cite{hessel2021clipscore} to measure the alignment between generated objects and provided text prompts, reporting the average CLIP scores across three scenes (\cref{fig:comparison}) in Table.~\ref{tab:clip_score}. Specifically, we randomly render 10 views for each scene and eliminate the background influences by cropping the bounding box region in the rendered image (\cref{fig:comparison}, first-row bottom right) for CLIP score calculation. From Table.~\ref{tab:clip_score}, the average CLIP score of our proposed method is 74.6, significantly outperforming other baselines by at least $20\%$.
\begin{table}[h]
    \centering
    \caption{\textbf{Quantitative comparison.} This table shows CLIP scores indicating the match between generated content and text prompts.}
    \label{tab:clip_score}
    \resizebox{1.0\columnwidth}{!}{
    \begin{tabular}{c|c|c|c}
    \hline
         & Blended-NeRF~\cite{gordon2023blended} & Spin-NeRF$^*$~\cite{mirzaei2023spin} & GO-NeRF\\
    \hline
    CLIP score~\cite{hessel2021clipscore} $\uparrow$ & 63.0 & 60.8 & \textbf{74.6}\\
    \hline
    \end{tabular}}
\end{table}

\subsection{Ablation Study}
\noindent \textbf{Inpainting SDS loss.} 
We conduct experiments using the vanilla SDS loss to validate the efficacy of inpainting SDS loss. From the results in \cref{fig:ab_all} (a), the inpainting SDS loss generates a cat with a complete body and proper pose, while the vanilla SDS only generates the head of a cat that does not seamlessly composite with the scene. In other words, the inpainting SDS loss can better utilize scene context to assist in 3D object generation.\\

\noindent \textbf{Saturation loss.}
In \cref{fig:ab_all} (b), we compare results with and without using saturation loss. When saturation loss is not applied, the generated objects, such as the backpack, exhibit an over-saturated appearance. To facilitate a clear comparison, we include visualized saturation maps at the bottom left, where bright regions indicate large saturation values. After regulating the object's saturation values during optimization, the generated objects coordinate well with the established scene.  
\\

\noindent \textbf{Sparsity loss and background augmentation.}
Simply compositing the generated object with the scene for optimization tends to introduce artifacts having a similar appearance to the scene background, as shown in \cref{fig:ab_all} (c) baseline results, where the green floaters are difficult to recognize in the scene with trees and grass as background. By introducing the sparsity loss that encourages the generated object to be compact, we observe a reduction in floaters, though they are still present  (\cref{fig:ab_all} (c) second column). To further eliminate floaters, we augment the background with pure white or black during training to make floaters apparent. Ultimately, the results are clean as displayed in \cref{fig:ab_all} (c) last column. The corresponding opacity maps are visualized in \cref{fig:ab_all} (c) second row, where floaters present as foggy patterns in the baseline results.\\



\noindent \textbf{Coarse-to-fine optimization.}
Instead of gradually increasing the hash grid resolution, we start training with a high-resolution hash grid and show results in \cref{fig:ab_all} (d) first column. The generated chair is inconsistent across different views because the capability of the high-resolution hash grid is too strong to overfit predefined camera views, potentially producing different chairs for different camera views. On the contrary, training using a coarse-to-fine scheme benefits the view consistency as shown in \cref{fig:ab_all} (d) last column. 

\subsection{Additional Experiments}
\noindent \textbf{Stereoscopic results for virtual reality.}
Our method focuses on the generation of virtual objects within 3D scenes, naturally supporting downstream virtual reality applications. Here, we render the 3D scenes into stereoscopic outputs, as displayed in \cref{fig:more_stereo} (a). To further visualize the stereo effects, we adopt the stereo transformer~\cite{li2021revisiting} to predict the disparity map from left and right view images. From \cref{fig:more_stereo} last column, the disparity map has distinguished foreground and background content, indicating good stereo effects. Please refer to our supplementary material for stereoscopic video results.\\

\noindent \textbf{3D reconstruction.}
In \cref{fig:more_stereo} (b), we display GO-NeRF's results rendered from different viewpoints and use these multi-view images to reconstruct a 3D Poisson mesh following Nerfstudio~\cite{tancik2023nerfstudio}. The mesh reconstruction is successful in \cref{fig:more_stereo} (b) right column, suggesting cross-view consistency, which can be further illustrated in the supplementary videos. The rendered novel-view images are better than the extracted mesh because NeRF excels at rendering high-fidelity images even if the underlying geometry is not perfect. \\

\noindent \textbf{Suspended objects.} To generate suspended objects, such as birds, we first create a 3D box using our interface in Sec.~\ref{sec: boudning_box} and additionally add a movement to the upwards direction (+z-axis) to obtain the suspended box, which is used to generate virtual objects. As displayed in \cref{fig:more_application} (a), our method successfully generates a bird suspended in the air.\\ 

\noindent \textbf{Multiple objects.} Our method allows for the gradual generation of multiple objects within a scene, either by producing them one by one or by combining pre-generated objects, thanks to the decomposed representation of both objects and the scene. As shown in \cref{fig:more_application} (b), the scene features several generated virtual objects, including a backpack, a traffic cone, and Pikachu.\\

\noindent \textbf{Style adaptation.}
In addition to text prompts, we also utilize reference images as conditions. As depicted in \cref{fig:more_application} (c) left, the reference image of a stone guides the generation of ``a stone chair" and ``a stone on the ground". The corresponding results are presented in the third and last columns of \cref{fig:more_application} (c). Compared to the results generated without a reference (shown in the second column of \cref{fig:more_application} (c)), the generated chair and stone more closely resemble the visual characteristics of the reference image.\\


\noindent \textbf{Scene editing.}
We demonstrate the capability of editing generated content in \cref{fig:more_application} (d). For instance, the appearance of the gray pigeon can be altered to white by fine-tuning with a modified text prompt. Furthermore, the separated representations of objects and scenes allow for the rearrangement of the bird by scaling and transforming the 3D bounding box used for object generation.\\


\noindent \textbf{Reflective surface.} 
We conduct experiments on more challenging scenarios, such as the reflective surface. Specifically, we tried to generate boats in the pond (\cref{fig:more_boat} left). Interestingly, the generated boats have reflections in \cref{fig:more_boat} right. One possible explanation is that diffusion priors, learned from large amounts of data, encourage reflections to be generated to promote the composition quality with the surrounding scene. Despite potential inaccuracies and incompetency of generated reflections, our findings demonstrate the possibility of exploiting diffusion priors to produce rendering-related effects. In the future, improving the physical correctness of the generation is worth exploring. 


\section{Conclusion}
This work has presented \emph{GO-NeRF}, a new method that advances the generation of text-controlled 3D objects directly within an established scene to craft new virtual environments. To achieve this, we offer users an intuitive interface to control generation positions and employ a compositional rendering formulation paired with tailored optimization objectives and training strategies for synthesizing 3D objects. 
Our methodology leverages diffusion priors from pretrained text-guided image inpainting models to facilitate the utilization of scene context and promote composition quality with the existing scene. 
Experimental results demonstrate the superiority of our approach across forward-facing and $360^o$ datasets. We envision our investigation will inspire future work endeavors in the domain of combining 3D reconstruction and generation for VR content creation, leveraging prior knowledge derived from extensive datasets. \\

\noindent \textbf{Limitations and future work.} 
The existing methods, such as SDS loss, aim at distilling 2D diffusion priors into 3D still exhibit a disparity with real-world applications.
By leveraging 2D diffusion priors, we have showcased that the generated content can align with the surrounding scene by adhering to scene conditions and producing rendering effects, such as shadows and reflections. 
Note that these effects, while indicative of coordination with the scene, may not yet reach a level of physical realism. 
Nevertheless, we note that the transfer of 2D diffusion priors, learned through data-driven manners, holds promise for enhancing realism in future renderings, including reflections, and shadows, by utilizing more advanced generation models and distilling techniques. 

Furthermore, the current single-box design for object generation can be improved. As illustrated in \cref{fig:more_boat}, incomplete and inaccurate reflections are observed due to the limitations of the predefined box structure in delineating the reflective space accurately. This discrepancy indicates that reflections may extend beyond the confines of the defined box. 
A possible solution to this issue could involve modeling both real and virtual objects using different boxes like the multi-space design as proposed in state-the-of-the-art work~\cite{yin2023multi}. 

Lastly, while our user-friendly interface for defining 3D boxes facilitates the generation of objects at specific locations, it lacks automation to some extent. Generating 3D objects without predefined locations necessitates a deeper understanding of scene information, which is beyond the scope and warrants further investigation.

\vspace{9pt}
\noindent \textbf{Supplementary material.} We also present three supporting demos/examples, for readers' information: 1) a video showcasing the generation process of our proposed approach alongside a comparison with other baselines; 2) an offline webpage presenting additional generated results (e.g., \cref{fig:more_cases}) in video format. This feature allows for easy exploration of the newly generated virtual environment from various perspectives; and 3) for users with access to a VR headset, the option to experience rendered stereoscopic videos offering both $360^o$ panoramic views and forward-facing scenes is available.

\bibliographystyle{abbrv-doi}

\bibliography{template}
\end{document}